\providecommand{\DIFdel}[1]{} % Don't show deleted text

\documentclass[letterpaper, 10 pt, conference]{ieeeconf}  % Comment this line out if you need a4paper

\IEEEoverridecommandlockouts                              % This command is only needed if 
\usepackage{graphicx}
\usepackage{amsmath}
\usepackage{amssymb}
\usepackage{booktabs}
 \usepackage{wrapfig}
 \usepackage[table]{xcolor}% http://ctan.org/pkg/xcolor
\usepackage{color,graphicx,soul}

\usepackage{multirow}
\usepackage{multicol}
\usepackage{balance}
\usepackage[ruled]{algorithm2e}

\soulregister{\cite}{7}

\title{\LARGE \bf
Targeted Hard Sample Synthesis Based on Estimated Pose and Occlusion Error for Improved Object Pose Estimation
}

\author{Alan Li and Angela P.~Schoellig% <-this % stops a space
\thanks{The authors are with the Dynamic Systems Lab, Institute for Aerospace Studies, University of Toronto, Canada, and affiliated with the Vector Institute for Artificial Intelligence. Angela P.~Schoellig is also with the Technical University of Munich (TUM). E-mails: { \{firstname.lastname\}@robotics.utias.utoronto.ca}} 
\thanks{This work was supported by resources from Epson Canada.}% <-this % stops a space
}

\begin{document}

\maketitle
\thispagestyle{empty}
\pagestyle{empty}

%%%%%%%%%%%%%%%%%%%%%%%%%%%%%%%%%%%%%%%%%%%%%%%%%%%%%%%%%%%%%%%%%%%%%%%%%%%%%%%%
\begin{abstract}

6D Object pose estimation is a fundamental component in robotics enabling efficient interaction with the environment. It is particularly challenging in bin-picking applications, where objects may be textureless and in difficult poses, and occlusion between objects of the same type may cause confusion even in well-trained models. We propose a novel method of hard example synthesis that is model-agnostic, using existing simulators and the modeling of pose error in both the camera-to-object viewsphere and occlusion space. Through evaluation of the model performance with respect to the distribution of object poses and occlusions, we discover regions of high error and generate realistic training samples to specifically target these regions. With our training approach, we demonstrate an improvement in correct detection rate of up to 20\% across several ROBI-dataset objects using state-of-the-art pose estimation models.

\end{abstract}

%%%%%%%%%%%%%%%%%%%%%%%%%%%%%%%%%%%%%%%%%%%%%%%%%%%%%%%%%%%%%%%%%%%%%%%%%%%%%%%%
\section{Introduction}

Object pose estimation, the task of determining the 3D position and orientation of an object within a scene, is fundamental to many computer vision applications. Its significance extends to autonomous navigation, object manipulation, and scene understanding. Over the years, various methods and algorithms have been developed to tackle this problem, including model-based, data-driven, and learning-based approaches.
Despite the progress, several challenges still impede the accurate and robust estimation of object poses, particularly in bin-picking, where repeated instances of the same part are populated in an unorganized manner within a bin, with various poses and various levels of occlusion. In recent years, convolutional neural networks (CNNs) have demonstrated great success in the problem of object pose estimation, particularly on datasets such as LINEMOD and T-LESS \cite{TLESS}, where objects are typically rich in texture and features, resulting in near-optimal color and depth images. However, single-view object pose estimation performs poorly in harder cases when data is missing, sparse, or noisy due to surfaces being glossy or transparent, or due to  occlusion and ambiguity from other objects \cite{goog}. In this paper, we explore the use of hard case mining, a technique rooted in machine learning and computer vision to mitigate these challenges and improve the reliability of object pose estimation systems.

\begin{figure}
  \begin{center}
    \includegraphics[width=\linewidth]{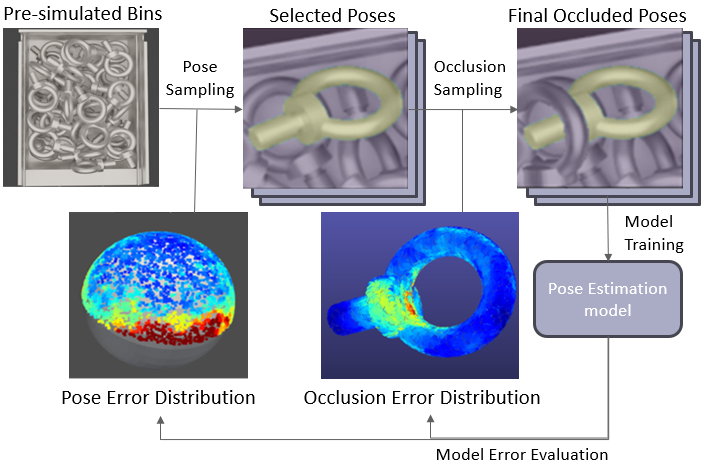}
    \end{center}
  \caption{The synthesis of hard samples using a set of pre-generated random bins. Poses are selected based on the error view-sphere, and realistic occlusions are added based on the occlusion model to generate the final training samples. After each training epoch, the model is evaluated again to update the error view-sphere and the occlusion model for the next round of hard sample synthesis.}
  \vspace{-3mm}
  \label{fig:multiview_heatmap}
\end{figure}

\begin{figure*}
  \begin{center}
    \includegraphics[width=\textwidth]{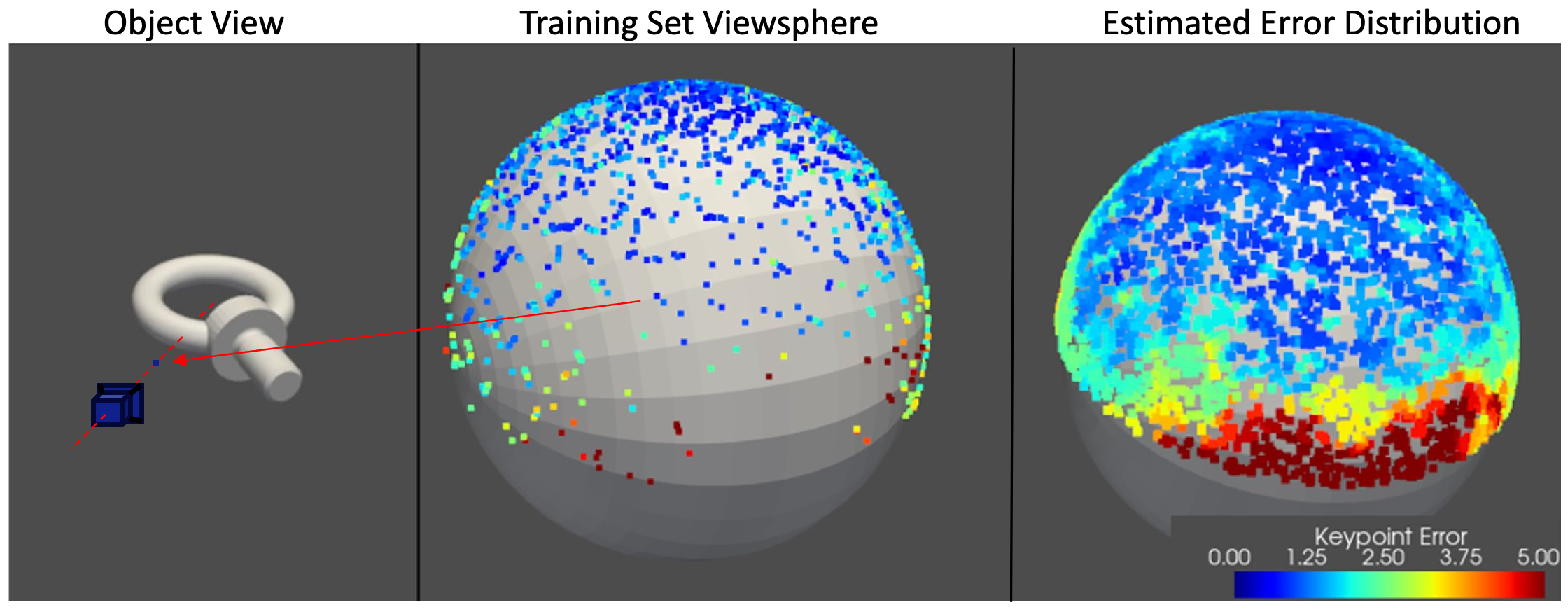}
    \end{center}
    \vspace{-4mm}
  \caption{Distribution of keypoint error across the view-sphere of Eye-bolt part. Each point on the sphere represents a single training sample where the camera is at a distance of approx. 500mm, and its viewpoint axis intersects with said point and the center of the object axis (left). The error distribution estimated from the training set error is shown on the right. Note that only half of the view-sphere is used due to symmetry of part.}
  \label{fig:viewdist}
\end{figure*}

Pose estimation of surrounding objects is a crucial yet difficult task for robotic systems that need to interact robustly with their environments. As the complexity of the environment and the potential interactions between objects grow, the required data to cover the distribution increases exponentially. Preparing comprehensive training data for 6D pose estimation is a formidable challenge. Traditional data generation methods often limit sampling of training data to only the object of interest. For instance, in SSD-6D \cite{kehl2017ssd6d}, a roughly uniform sphere of views is generated around the object and used to render the object against randomized backgrounds. This approach, although widely used, fails to model the potential interactions between the object and other foreground and background objects.

Another approach leverages physics simulators to generate realistic scenes of objects with natural household environments with realistic occlusions and interactions between objects and their backgrounds \cite{deepsim}. This was shown to produce higher overall performance, with the drawback that rarer poses and configurations are also underrepresented in the training data. These cases are typically harder cases which require more training samples rather than fewer to learn. Current approaches in pose estimation typically use a fixed training procedure and fixed training data, and are able to obtain decent overall performance, but are lacking in terms of reliability; certain arrangements of poses or occlusions may cause certain failure in these models \cite{deepsim}. 

We propose a novel approach in the context of object pose estimation which explores the input space of object poses and occlusions to sample additional novel training samples which specifically target these underrepresented and more difficult areas in the training data. The goal of this work is to demonstrate improvements to the reliability and overall accuracy of object pose estimation models compared to the state of the art, through three major technical innovations:

\begin{enumerate}
\item We develop a novel occlusion model from the training set which models the expected detection error given the parts of the object which are occluded, enabling the sampling of new occlusions based on error and likelihood. This is used to augment generated samples into hard cases to further train for specific types of occlusions.

\item We develop a method for synthesizing realistic new samples within a bin environment, given a desired object pose and occlusion areas, providing a way to generate novel hard cases in desired areas of the input pose space.

\item We demonstrate the effectiveness of our method using a continuous learning process where the pose error distribution is updated at each epoch and new samples are generated based on the updated error distribution, outperforming approaches with static training sets--converging up to 30\% faster and improving overall correct detection rate by up to 20\%. 
\end{enumerate}

\begin{figure*}
  \begin{center}
    \includegraphics[width=\textwidth]{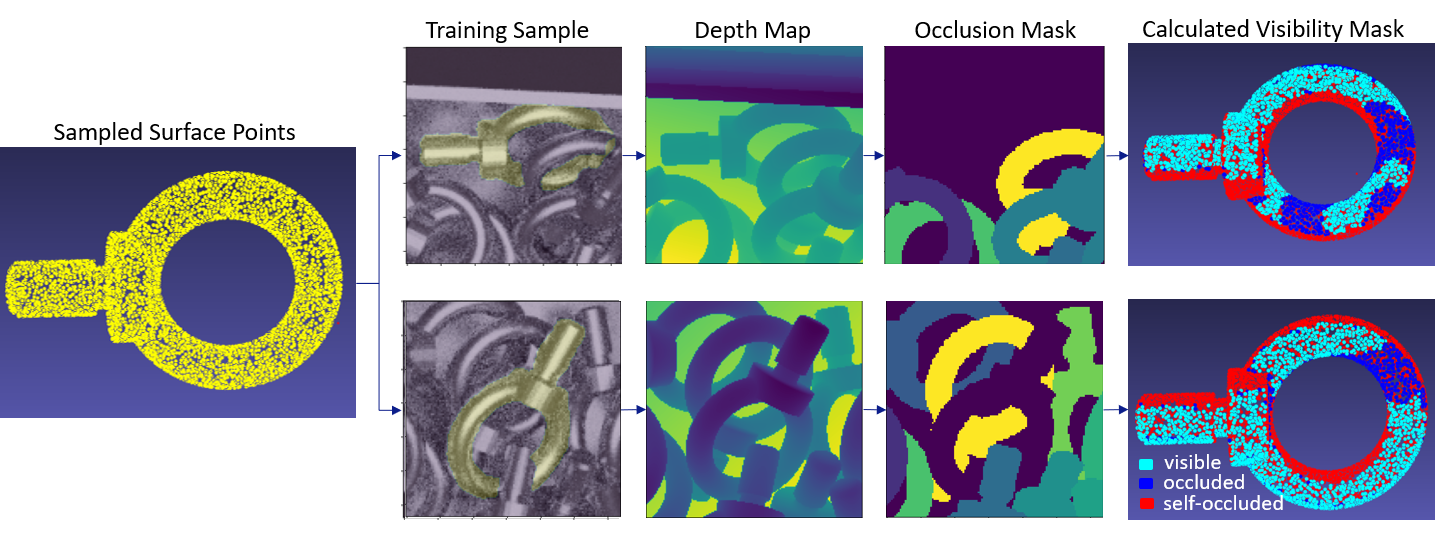}
    \end{center}
    \vspace{-4mm}
  \caption{An overview of the occlusion model generation, the goal of which is to estimate the expected error when a given point on the surface of the object is occluded. First a set of 5000 random points are sampled across the surface of the object, then pose estimator performance is evaluated for each training sample, the depth map is then used to determine the visibility of each of the sampled points, and the pose error is applied to points which should be visible based on the object pose but are occluded by other objects (blue).}
  \label{fig:occGen}
\end{figure*}

\begin{figure*}
  \begin{center}
    \includegraphics[width=\textwidth]{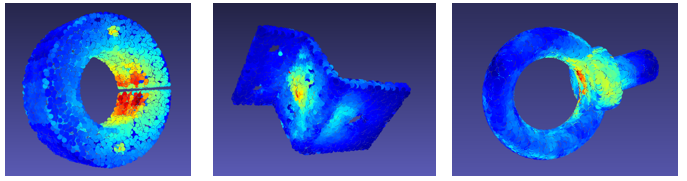}
    \end{center}
    \vspace{-4mm}
  \caption{Visualization of the estimated occlusion model for three parts, where redder areas represent higher expected error when occluded. Note that on the leftmost part, higher error is expected when the area around the slit is occluded, as the part is near-symmetric, and the visibility of the slit reduces the ambiguity of the pose. }
  \label{fig:occModel}
\end{figure*}

\section{Related Works}

Object pose estimation from images is a core problem in computer vision, with numerous applications in robotic manipulation, augmented reality, and autonomous driving. Methods such as hard negative mining, and hard sample exploration and synthesis are now only just being applied, finding growing success in the area of object pose estimation. 

\subsection{ Monocular 6D Object Pose Estimation }
Recent advances have leveraged the power of deep learning to surmount the shortcomings of traditional methods, which estimate pose based on local feature correspondences or template matching \cite{features}. These methods have been revisited using the advantages of CNNs to automatically learn features and implicitly estimate pose \cite{posecnn} \cite{pvnet}.  Many new methods leveraging architectures such as direct regression in GDR-Net \cite{gdrnet} and even transformers in FoundationPose \cite{foundationpose} have recently demonstrated their effectiveness in achieving state-of-the-art performance, though they typically follow similar processes for training data generation, and show limited focus towards improvement of these processes.

\subsection{ Hard Negative Mining in Object Detection }

Hard case mining, also known as hard negative mining, is a technique of augmenting training data with additional negative samples or samples with large training error with the goal of reducing error and false detections in these areas. It has recently proven effective for reducing errors in not only classification tasks, but regression tasks in computer vision, such as human pose estimation and face recognition \cite{schroff2015facenet}. Several approaches have been developed for automatically acquiring these hard case samples for training, including adversarial generation \cite{gong2021poseaug}, unsupervised data mining, and synthetic generation. In adversarial generation, an adversarial network is trained in a feedback loop with the discriminator to generate realistic hard samples with the goal of fooling the discriminator. These networks however, are notoriously hard to train due to the direct feedback loop, and often suffer from issues such as training divergence or mode collapse, where the generator ends up focusing down on only a few small areas of the input space \cite{modecollapse}. Unsupervised data mining can be used on large sets of unlabeled data to extract the hard cases found within. Typically this is done using a consistency metric or loss metric, such as detections that are isolated in time within a series of consecutive video frames \cite{vidsup}. This approach works well for object classification, as supervision can be clearly derived from negative samples, but in regression tasks like pose estimation with a continuous output, an incorrect pose estimate provides very limited information to discerning the correct estimate.

\subsection{ Exploration and Synthesis }
The generation of synthetic data to increase pose variation has found success in many pose estimation tasks, such as human pose \cite{chen2016synthesizing}, hand pose \cite{artiboost} and 6D object pose \cite{kehl2017ssd6d}, due to the difficulty of collection and manual annotation of real data.
In ArtiBoost \cite{artiboost}, active learning is used to explore and generate samples using a simulator across hand-object configuration and pose space. At each training epoch, the model performance on the training set is plotted across the space of all plausible configurations of a hand holding an object. This is used to sample new training points in the areas of higher error, which are then rendered synthetically and used to further tune the model at the next epoch. Our method takes a similar approach toward object pose estimation, but instead explores the space of all possible object poses and occlusions, while using a simulator to generate synthetic samples and an eye-in-hand camera to capture real samples.

\section{Method}

The purpose of our method is to synthesize realistic training samples with a frequency distribution proportional to the difficulty of the samples, as opposed to a uniform distribution or the natural distribution of samples within a bin. We assume a RGB-D image pipeline, along with a known object CAD model and simulator capable of generating realistic bin environments and synthetic RGB-D data for training.
The proposed approach, shown in Fig.~\ref{fig:multiview_heatmap}, is a novel training data generation pipeline, consisting of error estimation, synthesis, and update steps. The input pose estimation model may be pre-trained, or initialized from scratch. We evaluate the performance of the model across the distribution of possible poses and occlusions to determine the training data to be synthesized, and then further tune the model with a 50/50 split between the original random samples and the new targeted data. The tuned model is then re-evaluated to generate a new error distribution which is used in the generation of new samples for the following training epoch.

\subsection{Pose-space Error Distribution Estimation}

The goal of estimating the Pose-space error distribution is to enable rapid sampling of new object views with likelihood weighted by their expected pose error, with higher error views occurring more frequently than lower error views. This is accomplished through the function $P(\theta,\phi)$, which outputs the expected model error for a given out-of-plane rotation parameterized by the polar and azimuthal angles $(\theta,\phi)$. We focus the model on the out-of-plane rotation axes, as in-plane samples may be generated simply by rotating the image samples. We also assume a large enough camera distance of approximately 500mm, also matching the mean distance to camera in the real ROBI dataset, to minimize the projective effects of small translations on the appearance of the object.

The expectation of model error output by $P(\theta,\phi)$ is taken over the training set samples present over a small neighborhood of the input angles $(\theta,\phi)$. Specifically, we evaluate the network performance on the full training set and plot the estimation error of each sample w.r.t. the out-of-plane rotation $(\theta,\phi)$ as a point on a unit sphere, as seen in Fig.~\ref{fig:viewdist} (middle). As our chosen network minimizes mainly keypoint error in its loss function, we use this to represent the model estimation error. We use the unit sphere training points to build a KD-Tree for nearest neighbors search for a given input angle. The expected model error for the input angle $(\theta,\phi)$ is then estimated by taking the mean model error (keypoint error in our selected model, but pose error metrics such as ADD may be used instead with minimal difference) of the five nearest neighbors $nn_{err,5}$, along with a penalty term $nn_{dist,5}$ proportional to the mean distance of the five nearest neighbors. The purpose of this penalty is to encourage a balance between exploration of areas with fewer training samples, and exploitation of areas with known higher error, similar to the approach outlined in ArtiBoost \cite{artiboost} for estimation of model error hand-object space.

In practice, we find that taking the maximum between the nearest neighbors estimate $nn_{err,5}$ and the mean euclidean distance to the nearest neighbors $nn_{dist,5}$ multiplied by a scaling factor $\beta = 5$,
\begin{equation}
P(\theta,\phi) = max(nn_{err,5}(\theta,\phi),\beta nn_{dist,5}(\theta,\phi)) ,\
\end{equation}

 offers a smooth transition from exploration of sparse areas to exploitation of higher error poses, as each new generated sample is added to the training set in the next training epoch. Setting $\beta = 5$ is equivalent to scaling the unit sphere to a sphere of radius 5, and evaluating the nearest neighbors and their distances on this new sphere.

\subsection{Occlusion-space Error Distribution Estimation}

Occlusion randomization in training approaches typically involves dropout/CutOut \cite{cutout} of random areas in the sample, or a random 2D pasting of another object on top of the sample. While computationally fast and easy to implement, it does not accurately simulate the 3D effects of a real occluding objects, particularly the effect on the depth map and lighting of the object. 
We propose a novel method of modeling the effect of specific occlusions on network performance by estimating the expected keypoint error of a sample object $i$ when a given point on its surface is occluded by another object $j$. We use the keypoint error on the training set to represent network performance for our model, but overall pose error may be used as well. This produces an error distribution across the surface of the object, shown in Fig.~\ref{fig:occModel}, which we may use to guide our occlusion-aware hard case synthesis.
The overall approach involves processing each sample in the training set, and determining which points on the object should be visible based on the camera viewpoint but are blocked by another occluding object. The relative object poses of occluding objects are recorded and the occluded points are then updated to reflect the observed network performance on the given sample:

\begin{figure*}
  \begin{center}
    \includegraphics[width=\textwidth]{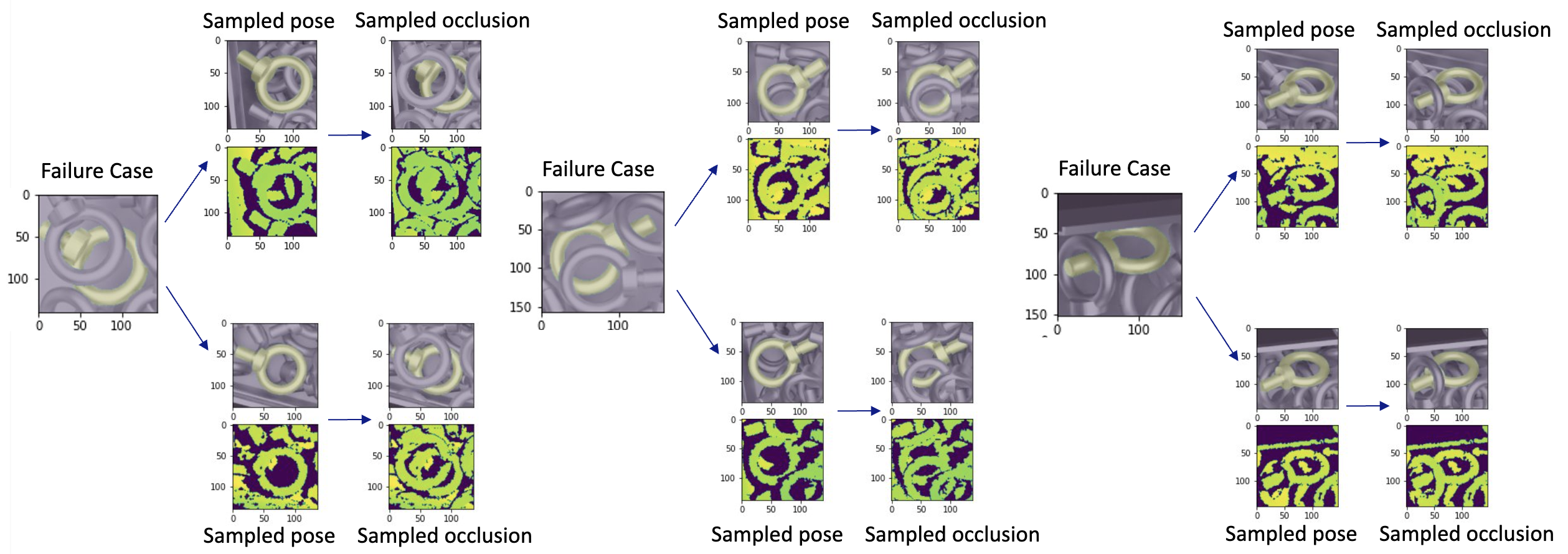}
    \end{center}
    \vspace{-4mm}
  \caption{Three examples of the hard case sample generation pipeline, starting with the original nearest neighboring failure case from the sampled point in the pose error distribution, to the pose sampled from the pre-generated bins, to the sampled occluding objects added and rendered into the scene. Note the realistic occlusions and object arrangements in both RGB and depth images achieved with our method.}
  \label{fig:genData}
\end{figure*}

\begin{enumerate}

    \item Sample 5000 points $x_k, \, k \in  \{ 1...5000\}$ uniformly across the surface of the object model, as seen in the first image of Fig.~\ref{fig:occGen}.
    
    \item Let $(u^i_{k},v^i_{k})$ be the pixel coordinates of the $k^{th}$ sample point $x_k$, on the object $i$ in the scene. For each image containing object ${i}$, project each point $x_k$ into pixel coordinates  $(u^i_{k},v^i_{k})$  of the image using the ground truth camera-to-object pose  $\boldsymbol {T}^{c}_{i}$ and camera intrinsic matrix $\boldsymbol K$, 
    
    \begin{equation}
\begin{bmatrix}
 u^i_{k}\\
 v^i_{k}\\
 1
 \end{bmatrix} =\boldsymbol K  \dfrac{1}{z^{c_i}_k}\boldsymbol{T}^{c}_{i}\boldsymbol{x}_k , 
\end{equation}
    where $z^{c_i}_k$ is the z coordinate of the sample point $x_k$ on object $i$ in the camera frame. 
    
    \item Determine the occlusion $V_k(i,j)$ of each sample point $x_k$ on object $i$ by neighboring object $j$ in the scene using the depth map of the scene $d(u,v)$, the z coordinate of the sample point in the camera frame $z^{c_i}_k$, and the object mask function $m(j,u,v)$, which returns 1 if the pixel at $(u,v)$ is occupied by object $j$ and 0 otherwise:  
\begin{equation}
V_k(i,j) = (z^{c_i}_k - d(u^i_k,v^i_k))  *  m(j,u^i_k,v^i_k) ,\
\end{equation}
    where sample point $\boldsymbol{x}_k$ on object ${i}$ is occluded by the object $j$ only when $V_k(i,j)$ is a positive nonzero number.
        
    \item Let the training set $\mathcal{D}$ consist of the set of tuples $(i,j)$ in which the occluding object $j$ appears in an image with the sample object $i$. Let $\boldsymbol{T}^i_j$ be the relative pose between the occluding object $j$ and the sample object $i$, and let $\mathcal{T}_k$ be the set of all relative poses present in the training data $\mathcal{D}$ which result in the occlusion of the sample point $x_k$. For all combinations of occluding object $j$ and sample object $i$ present in the training data, we add the relative pose $\boldsymbol{T}^i_j$ to the set $\mathcal{T}_k$ if the occlusion $V_k(i,j)$ is greater than zero:
    \begin{equation}
        \mathcal{T}_k = \{\boldsymbol{T}_{j}^{i}  \,| \, V_k(i,j) > 0  \,\,\forall \,(i,j) \in \mathcal{D} \} .
    \end{equation}

        \item Let $l(i)$ be the model estimation error of sample object $i$. We can now also calculate the expected pose error $L(k)$ when sample point $x_k$ is occluded, by averaging the estimation error $l(i)$ for each sample object $i$ in the poses contained the set $\mathcal{T}_k$:
    \begin{equation}
        L(k) = \frac{1}{N} \sum_{i \in I} l(i) ,  \, I = \{i \,|\, \boldsymbol{T}^i \in \mathcal{T}_k \} ,
    \end{equation}
        where N is the total number of elements in $\mathcal{T}_k$.
 \end{enumerate}

From equations $(4)$ and $(5)$, we now have a set of poses of neighboring objects which occlude any given sample point $x_k$, but also the expected pose error when the sample point $x_k$ is occluded.

\begin{figure*}
  \begin{center}
    \includegraphics[width=\textwidth]{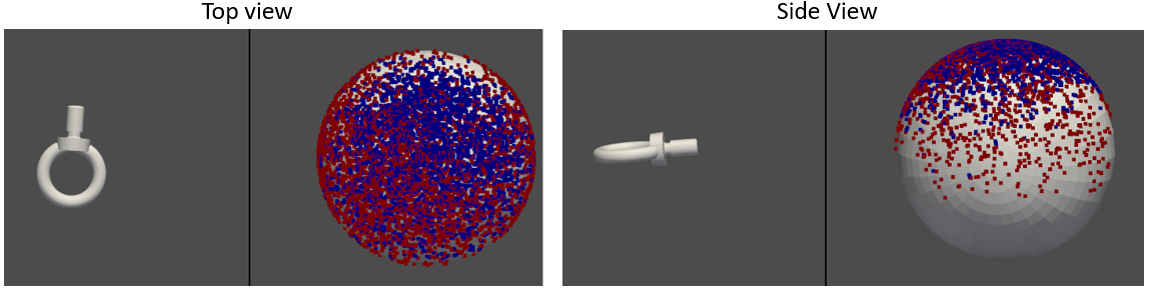}
    \end{center}
    \vspace{-4mm}
  \caption{Plots of the pose distribution of training data generated using random bins (blue) versus targeted training data generation with our method (red). Our method achieves much higher coverage of the less common side views of the object.}
  \label{fig:randVsHard}
\end{figure*}

\subsection{Hard Case Sample Generation}

With the Pose-space and Occlusion-space error distributions, we can sample new hard cases based on the error likelihood, and render them in realistic bin-picking backgrounds using a set of pre-simulated bins. The approach involves first sampling an object pose $\boldsymbol{T}^c_m$ using the Pose-space error distribution $P(\theta,\phi)$, conditioning the occlusion-space error distribution $L(k)$ given the visible points $\mathcal{V}_m$ from the sampled pose, and then sampling from the conditioned occlusion-space distribution $P(k\,|\,k\in\mathcal{V}_m)$: 

\begin{enumerate}

\item Let $m$ be an object in the set $\mathcal{M}$ of new training samples we would like to generate. Sample the polar and azimuthal angle for the object pose $\boldsymbol{T}^c_m$ for sample $m$ from the Pose-space distribution in equation $(1)$:
\begin{equation}
    (\theta_m,\phi_m) \sim P(\theta,\phi)
\end{equation}

\item From the set of pre-simulated scenes, find an object $o$ with camera-to-object pose $\boldsymbol{T}^c_o$ within 5 degrees of the sampled angles from the previous step, and transform all objects in the scene with the pose difference $\boldsymbol{T}^o_m$ between $\boldsymbol{T}^c_o$ and ($\theta_m,\phi_m$). This effectively moves the camera to the desired angles from a nearby pose, and creates $\boldsymbol{T}^c_m$:
\begin{equation}
    \boldsymbol{T}^c_m = \boldsymbol{T}^c_o \boldsymbol{T}^o_m
\end{equation}

\item Find the set of visible sample points $\mathcal{V}_m$ in the pose $\boldsymbol{T}^c_m$ by rendering the depth map with the object pose $d(u,v)$, calculating the pixel coordinates of each point $(u^m_{k},v^m_{k})$, and performing depth testing, similarly to equation $(3)$:
\begin{equation}
    \mathcal{V}_m = \{k \,|\, 
     z^{c_m}_k - d(u^m_k,v^m_k) > 0\}, 
\end{equation}
where we do not use the mask term $m$ as we are looking only for points which are not self-occluded by the object itself.

\item From the visible points $\mathcal{V}_m$, calculate the distribution $P(k\,|\,k\in\mathcal{V}_m)$ using $L(k)$ from Eq. 5, representing the distribution of estimation error of occlusions for the given sampled pose. Then sample a point $k_m$ from this distribution to be occluded:

\begin{equation}
    k_m \sim P(k\, |\, k \in \mathcal{V}_m) = \dfrac{L(k, k \in \mathcal{V}_m)}{\sum\limits_{j\in\mathcal{V}_m}(L(j))}
\end{equation}

\item We now sample a neighboring object pose $\boldsymbol{T}^m_n$ uniformly from the set of occluding poses $\mathcal{T}_{k_m}$ which result in occlusion of sample $k_m$, as defined in equation $(4)$: 
\begin{equation}
    \boldsymbol{T}^m_n \sim  {U}(\boldsymbol{T}\in\mathcal{T}_{k_m}) .
\end{equation}

where $U$ is the uniform distribution across the set $\mathcal{T}_{k_m}$.

\item Add a rotation perturbation of $\pm2$ degrees in to create $\boldsymbol{T}^m_{n\prime}$ and transform the neighboring object pose to camera coordinates $\boldsymbol{T}^c_n$ for rendering: 
\begin{equation}
    \boldsymbol{T}^c_n =\boldsymbol{T}^c_m\boldsymbol{T}^m_{n\prime}
\end{equation}

\item Add the occluding object pose $\boldsymbol{T}^c_n$ to the scene and render the final RGB and depth images for the sample $m$ with remaining scene objects in the background.

\end{enumerate}

Fig.~\ref{fig:genData} shows three examples of the hard case sample generation pipeline, starting from the original nearest neighboring failure case from the sampled point in the pose error distribution, to the pose sampled from the pre-generated bins, to the sampled occluding objects added and rendered into the scene.

\begin{figure*}
  \begin{center}
    \includegraphics[width=\textwidth]{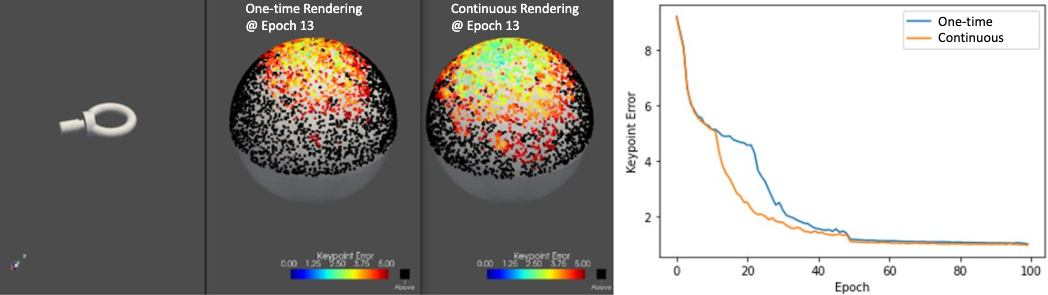}
    \end{center}
    \vspace{-4mm}
  \caption{Continuous updating of the pose error and online data generation at each training epoch results in faster convergence and an overall lower error of $\sim$8\% at epoch 100, compared to a single training data generation step. The test set error distribution at training epoch 13 is shown on the left figure, for both the regular one-time rendering and the continuous rendering.  }
  \label{fig:onlineTraining}
\end{figure*}

\subsection{Online Training}
The described exploration and synthesis steps are performed during the training of the model; that is, at each training epoch, the pose and occlusion models are re-evaluated using the performance of the latest model, and new samples are generated for the next training epoch. The distribution of the validation data is not changed, as we are interested in improving the overall performance of the network and not only the hard cases. Fig.~\ref{fig:onlineTraining} demonstrates that this online training process gives faster training convergence and overall lower loss at a given training epoch compared to a single error evaluation and data generation step.

\subsection{Single-view Keypoint and Heatmap Estimation } The keypoint detection network is based off PVNet, a pixel-wise voting network for 6D pose estimation of a known object with a CAD model\cite{pvnet}. The network takes as input an image patch containing an object of interest, and predicts a keypoint location from each pixel input, in the form of an offset to the pixel location itself. The training data is generated synthetically by simulating the random arrangement of parts within a bin, followed by realistic rendering of the scene to RGB in Blender. The depth images are rendered using NxView, a depth camera simulator supplied with the Ensenso camera. Further details of the network can be found in our previous work. \cite{multiviewKP}, though our proposed approach is agnostic to the pose estimation network, and we demonstrate its effectiveness with the GDRNPP \cite{gdrnpp} model as well.

\section{Experiments}
The effectiveness of the hard case generation is shown using the ROBI dataset. This dataset includes a total of 63 bin-picking scenes captured with an active stereo depth Ensenso N35 sensor \cite{robi}. For each scene, a view sphere totalling 88 RGB images and depth maps are captured from an Ensenso N35 stereo depth camera, and are annotated with accurate 6D poses of
visible objects and an associated visibility score. A total of seven different objects are found in the dataset, arranged randomly within bins. The objects are metallic and highly reflective, with varying levels of symmetry. This results in a dataset representing the most difficult scenes typically found in bin picking applications, with self-occlusions between multiple instances of the same object, and reflections causing missing depth data and false RGB edges.
\begin{table*}[h]
\caption{Evaluation and Comparison Results on the ROBI Dataset}
\label{table_examp}
\begin{center}
\begin{tabular}{c||c c c c c c}

\multirow{2}{4em}{Object} & \multicolumn{6}{c}{ Correct Detection Rate ($<$10\% ADD-S Error) }  \\
%\cline{2-11}
  & DC-Net &MP-AAE & Sim2Real & FoundationPose & Ours (Baseline) & Ours (Pose+Occ.)  \\
\hline
Zigzag  & 30.9 & 92.4 & 86.7 & \textbf{97.1} & 89.3 & 95.5 \\ 
%\hline
Eye Bolt  & 53.5& 67.3 & 89.2 & 93.9 & 82.5 & \textbf{96.7}\\
%\hline
DIN  & 18.7& 21.8 & 24.2 & \textbf{64.3} & 31.8 & 56.4 \\
%\hline
Gear &77.6& 82.1 & 76.8 & \textbf{87.4} & 72.7 & 81.9  \\
%\hline
Tube Fit. & 74.8 & 91.6 & \textbf{97.5} & 94.2 & 71.1 & 89.3 \\
%\hline
Screw  & 67.7& 90.3 & 78.1 & 76.4 & 89.6 & \textbf{93.8}\\
%\hline
D-Sub &10.6&7.35 & 14.6 & 47.4 & 25.6 & \textbf{48.2} \\
\hline
\end{tabular}
\end{center}
\vspace{-4mm}
\end{table*}
Table 1 shows the pose estimation results on several objects in the ROBI dataset. The ROBI test set consists of the top layer of objects which are less than 40\% occluded in the majority of the chosen views. The baseline performance is measured using several state-of-the-art approaches, including DC-Net \cite{dcnet}, MP-AAE \cite{aae}, Sim2real \cite{sim2real}, and FoundationPose \cite{foundationpose}. DC-Net and MP-AAE are trained using randomized generation of synthetic images, while Sim2real also includes the addition of pseudo-labeled real samples. FoundationPose is trained to be object-agnostic, with the object model instead input or reconstructed at test time. The base keypoint network is trained using a synthetic dataset of 15000 samples generated with random simulated object poses within bins, while the augmented network uses 8000 of these random samples and 7000 targeted synthetic samples, which are generated during the training process itself. This means that the network begins with zero targeted samples at the first epoch of training, eventually reaching 7000 samples near the last epochs of training. The training time is increased by approximately 50\%, from 8 hours to around 12 hours on an NVIDIA Titan V, though many parallelization optimizations have yet to be applied. The correct detection rate is defined as the percentage of visible objects in the bin associated with an output pose less than 10\% ADD-S error, as proposed by Hinterstoisser et al. \cite{hinter}. The average distance between vertices on the nearest symmetrical ground truth pose of the model and the corresponding vertices on the predicted pose is calculated, and predictions with less than 10\% average error w.r.t. the object diameter are considered correct detections.

\begin{table}[h]
\caption{Ablation Study on Training Augmentation in T-LESS using GDRNPP as base model}
\label{table_example2}
\begin{center}
\begin{tabular}{c||c c c }
\hline
\multirow{2}{6em}{T-LESS Scene} & \multicolumn{3}{c}{Detection Rate (ADD-S, $AR_{MSSD}$)}  \\
%\cline{2-11}
  & Baseline & Ours (Pose)  & Ours (Pose+Occ.)  \\
\hline
All  & (89.1, 81.3) & (92.4, 83.0)  & \textbf{(92.9, 83.3)} \\ 
%\hline
 Bin-Picking Only  & (80.2, 74.2) & (88.6, 80.9)  & \textbf{(91.2, 82.7)}\\
%\hline

\hline
\end{tabular}
\end{center}
\vspace{-4mm}
\end{table}

Table 2 presents an ablation study using the more widely known T-LESS dataset \cite{TLESS}. We show results on both the full dataset, and a subset containing only scene 20, the only bin-picking scene with objects presented in higher difficulty poses with more intra-class occlusion. We demonstrate the model-agnostic nature of our approach by applying it to GDRNPP \cite{gdrnpp}, winner of the BOP 2022 challenge. The baseline training approach from GDRNPP uses a total of 15K training samples, split between a set of uniformly sampled viewpoints and with randomized views with domain-randomized occluding objects and backgrounds. We demonstrate the effectiveness of our approach through the results of pose and pose-and-occlusion targeted training data generation, normalized to use the same total number of training samples as the baseline, closing the large $\sim$10\% gap in ADD-S and $AR_{MSSD}$ \cite{bop2020} between the bin-picking and standard scenes, but also demonstrating a modest improvement in the full dataset as well.

\begin{figure}
    \includegraphics[width=\linewidth]{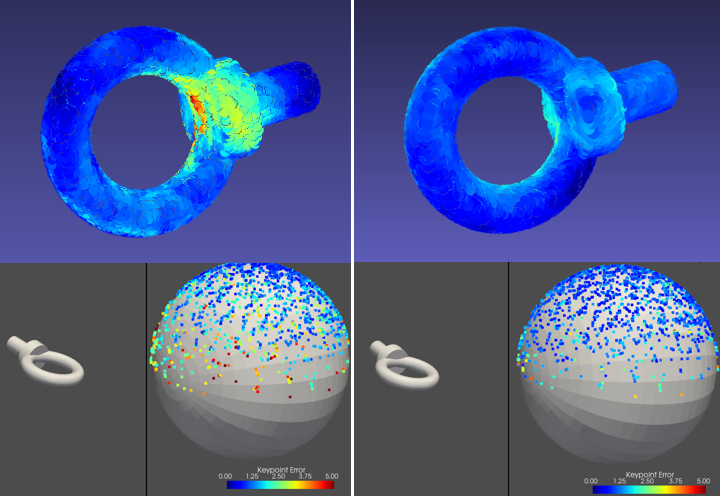}
    \vspace{-4mm}
  \caption{Keypoint error in view and occlusion space of original model (left) and model trained with our method (right). }
  \label{fig:beforeafter}
\end{figure}

\begin{figure}
    \includegraphics[width=\linewidth]{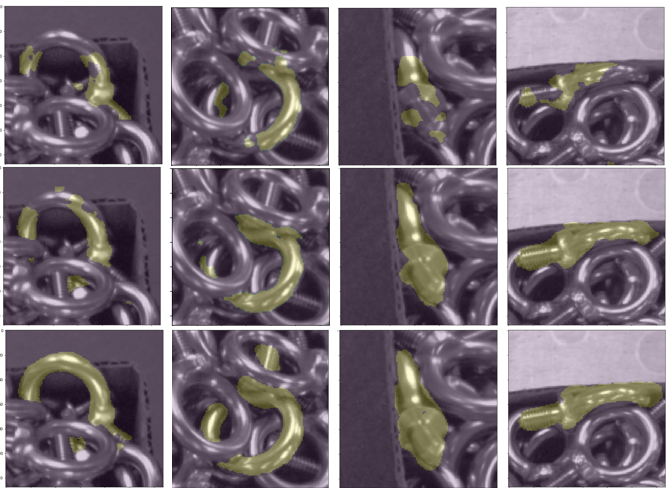}
    \vspace{-4mm}
  \caption{Segmentation performance of base model (top) compared with model trained using pose sampling (middle) and model train using pose and occlusion sampling (bottom). Note the improvement in object instance persistence when adding occlusion sampling.}
  \label{fig:posevsseg}
\end{figure}

\section{Discussion and Limitations}

Our training method turns gives our baseline model state-of-the-art performance competitive with FoundationPose, outperforming it in Eye Bolt, Screw and D-Sub, where there are larger differences in domain between the object models and the real object. 
The results also showed a large reduction in failure cases not only on the synthetic validation data, but also the test set in the real domain. Fig. \ref{fig:beforeafter} shows the reduction in estimation error on the Eye Bolt part in both occlusion and pose space on the real test data. We demonstrate not only that our method is effective in reducing the failure cases, but also that these failure cases occur across domains, and a large portion are fixed in the real domain simply by training on the synthetic domain equivalents. Fig.~\ref{fig:randVsHard} demonstrates the difference in distribution between training data which is generated from randomized bins of simulated parts, versus the distribution of the poses generated using our method. Notice that the random distribution is centered around the stable poses, where the object is lying flat, whereas our method generates more poses around the side views, which are typically rarer but also more difficult poses for the network to estimate. The importance of the occlusion sampling can be seen in Fig. ~\ref{fig:posevsseg}, where the network is able to fully segment the eye-bolt object under difficult occlusions and even when is it bisected by another object, whereas the original model and model trained only with pose sampling are unable to do so. 

In the Tube Fitting and Gear objects, the original PVNet-based network itself performs notably worse, due likely to some drawbacks on how the network handles high orders of symmetry, but our method still demonstrates a large jump in performance.

Although we were unable to test our approach on FoundationPose or GPose2023, the current leader on the BOP challenge, as they have yet to release their training code,  we expect similar performance gains in bin-picking scenes, as demonstrated in our experiments with GDRNPP as the base model, due to their similar baseline data generation techniques of uniform and random pose sampling. 

We acknowledge that our evaluation is targeted towards bin-picking scenes of textureless objects; when evaluating over all T-LESS scenes, the improvement from our approach is a more modest 3.8\%. This is likely due to the other scenes having easier, more common poses, less occlusion, and more diverse and textured objects, which may benefit less from our training approach.

\balance

\section{Conclusion}

The main contribution of this paper is a novel method of boosting the training of pose estimation models on difficult textureless objects through modeling of the relationship between the object pose, occlusions, and estimation error, and the targeted generation of new training data. We demonstrate an improvement in CDR with our training method of up to 20\% on the ROBI dataset, and 10\% on the T-LESS bin-picking scene, without increasing the number of training samples required.

%%%%%%%%%%%%%%%%%%%%%%%%%%%%%%%%%%%%%%%%%%%%%%%%%%%%%%%%%%%%%%%%%%%%%%%%%%%%%%%%

%%%%%%%%%%%%%%%%%%%%%%%%%%%%%%%%%%%%%%%%%%%%%%%%%%%%%%%%%%%%%%%%%%%%%%%%%%%%%%%%

%%%%%%%%%%%%%%%%%%%%%%%%%%%%%%%%%%%%%%%%%%%%%%%%%%%%%%%%%%%%%%%%%%%%%%%%%%%%%%%%

%%%%%%%%%%%%%%%%%%%%%%%%%%%%%%%%%%%%%%%%%%%%%%%%%%%%%%%%%%%%%%%%%%%%%%%%%%%%%%%%

\end{document}